\documentclass[]{antgroup}
\PassOptionsToPackage{numbers, compress}{natbib}
\usepackage{antgroup}

\usepackage{graphicx}
\usepackage{tikz}
\usepackage{todonotes}
\usepackage{multirow}
\usepackage{amsmath}
\usepackage{cleveref}
\usepackage{subcaption}
\usepackage{multicol}
\usepackage{amssymb}
\usepackage{array}
\usepackage{bm}
\usepackage{enumitem}
\usepackage{algorithm}
\usepackage{algpseudocode}
\usepackage{tabularx}
\usepackage{booktabs} 
\usepackage{bbm}
\usepackage{makecell}
\usepackage{wrapfig}
\usepackage{colortbl}
\usepackage{threeparttable}
\usepackage{fontawesome5}
\usepackage[normalem]{ulem}
\usepackage{enumitem}
\usepackage{caption}
\useunder{\uline}{\ul}{}

\usepackage{amsmath,amsfonts,bm}

\def\eqref#1{equation~\ref{#1}}

\def\1{\bm{1}}

\DeclareMathAlphabet{\mathsfit}{\encodingdefault}{\sfdefault}{m}{sl}
\SetMathAlphabet{\mathsfit}{bold}{\encodingdefault}{\sfdefault}{bx}{n}

\newcommand*\justify{%
  \fontdimen2\font=0.4em
  \fontdimen3\font=0.2em
  \fontdimen4\font=0.1em
  \fontdimen7\font=0.1em
  \hyphenchar\font=`\-
}

\renewcommand{\texttt}[1]{%
  \begingroup
  \ttfamily
  \begingroup\lccode`~=`/\lowercase{\endgroup\def~}{/\discretionary{}{}{}}%
  \begingroup\lccode`~=`[\lowercase{\endgroup\def~}{[\discretionary{}{}{}}%
  \begingroup\lccode`~=`.\lowercase{\endgroup\def~}{.\discretionary{}{}{}}%
  \catcode`/=\active\catcode`[=\active\catcode`.=\active
  \justify\scantokens{#1\noexpand}%
  \endgroup
}

\usepackage{amsmath}
\usepackage{amssymb}
\usepackage{amsfonts}                               
\usepackage{amsthm}
\usepackage[mathcal]{eucal}
\usepackage{mathrsfs}
\usepackage{bm}                                     
\usepackage{blkarray}                               
\usepackage{nicefrac}                               

\usepackage{wrapfig}
\usepackage{graphicx}                               
\renewcommand\thesubfigure{(\alph{subfigure})}      
\usepackage{caption}
\usepackage{cleveref}

\usepackage{tikz}                                          
\usepackage{circuitikz}
\usetikzlibrary{patterns,snakes}
\usetikzlibrary{positioning,calc,fit,decorations.pathmorphing,shapes.geometric, shapes.gates.logic.US, calc}
\usetikzlibrary{arrows,arrows.meta,decorations.markings,shapes,shapes.arrows}
\usetikzlibrary{decorations,decorations.pathreplacing}
\usetikzlibrary{backgrounds}
\usepackage{filecontents}                           
\usepackage{pgfplots}
\usepackage{pgfplotstable}
\usepgfplotslibrary{groupplots}
\usepackage{scalefnt}
\pgfplotsset{compat=newest}

\usepackage{xcolor}

\definecolor{LightGray}{gray}{0.9}

\definecolor{myPink}{HTML}{EE6B98}
\definecolor{myBlue}{HTML}{6BB8FA}
\definecolor{myPurple}{HTML}{9F63F0}
\definecolor{myBlueBase}{HTML}{147BFF}

\renewcommand\thefootnote{}

\title{LLaDA2.0: Scaling Up Diffusion Language Models to 100B}

\author{\centering
Tiwei Bie$^1$,
Maosong Cao$^1$,
Kun Chen$^1$,
Lun Du$^1$,
Mingliang Gong$^1$,
Zhuochen Gong$^1$,
Yanmei Gu$^1$,
Jiaqi Hu$^{1,3}$,
Zenan Huang$^1$,
Zhenzhong Lan$^{1,4,\dag}$,
Chengxi Li$^1$,
Chongxuan Li$^2$,
Jianguo Li$^{1,\dag}$,
Zehuan Li$^1$,
Huabin Liu$^1$,
Lin Liu$^1$,
Guoshan Lu$^1$,
Xiaocheng Lu$^{1,5}$,
Yuxin Ma$^1$,
Jianfeng Tan$^1$,
Lanning Wei$^1$,
\\
Ji-Rong Wen$^2$,
Yipeng Xing$^1$,
Xiaolu Zhang$^1$,
Junbo Zhao$^{1,3,\dag}$,
Da Zheng$^{1,\dag}$,
Jun Zhou$^1$,
Junlin Zhou$^1$,
\\
Zhanchao Zhou$^{1,4}$,
Liwang Zhu$^1$,
Yihong Zhuang$^1$
}

\footnotetext{Authors are listed in alphabetical order based on last name. $\dag$ indicates tech-leaders.}

\affiliation{$^1$Ant Group, $^2$Renmin University of China,  $^3$Zhejiang University,\\$^4$Westlake University, $^5$HongKong University of Science and Technology}

\begin{document}
\maketitle

\renewcommand\thefootnote{\arabic{footnote}}
\setcounter{footnote}{0}

\begin{abstract}

This paper presents LLaDA2.0 --- a tuple of discrete diffusion large language models (dLLM) scaling up to 100B total parameters through systematic conversion from auto-regressive (AR) models --- establishing a new paradigm for frontier-scale deployment. 
Instead of costly training from scratch, 
LLaDA2.0 upholds knowledge inheritance, progressive adaption and efficiency-aware design principle, and seamless converts a pre-trained AR model into dLLM with a novel 3-phase block-level WSD based training scheme: progressive increasing block-size in block diffusion (warm-up), large-scale full-sequence diffusion (stable) and reverting back to compact-size block diffusion (decay).
Along with post-training alignment with SFT and DPO, we obtain LLaDA2.0-mini (16B) and LLaDA2.0-flash (100B), two instruction-tuned Mixture-of-Experts (MoE) variants optimized for practical deployment. By preserving the advantages of parallel decoding, these models deliver superior performance and efficiency at the frontier scale.
Both models were open-sourced.\\

\textbf{Huggingface:} \url{https://hf.co/collections/inclusionAI/llada-20}

\end{abstract}

\begin{figure}[hb]
    \centering
    \small
    \includegraphics[width=0.95\linewidth]{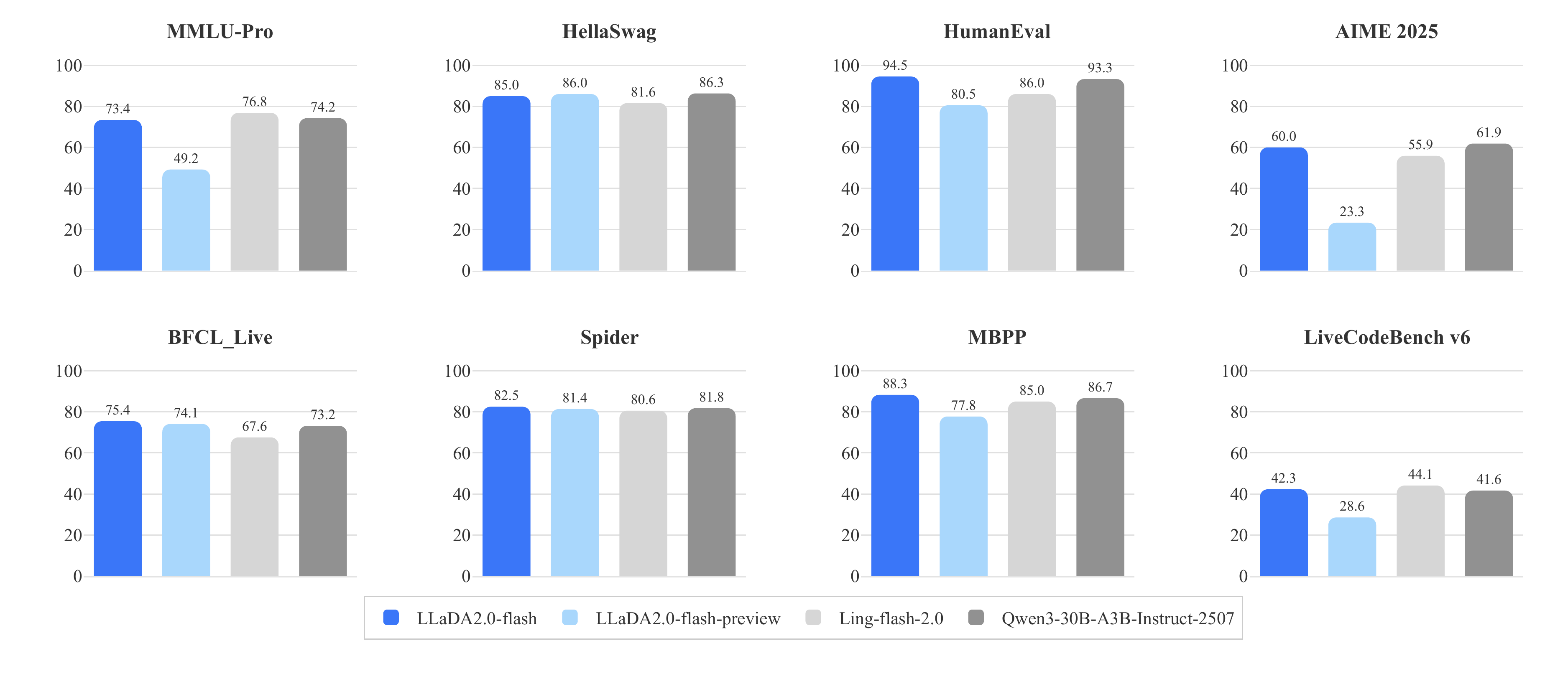}
    \vspace{-2ex}
\caption{\texttt{LLaDA2.0-flash} main results.}\label{fig:llada2_flash_main_bench}
\vspace{-2ex}
\end{figure}

\newpage
\section{Introduction}

Large Language Models have achieved remarkable success through the AR paradigm, modeling sequences via next-token prediction with strict left-to-right causal dependencies~\citep{hurst2024gpt4o,grattafiori2024llama3,yang2025qwen3}. This approach naturally aligns with the sequential structure of language and enables efficient training through next-token likelihood maximization. However, the very success of this paradigm creates fundamental limitations: the sequential generation process imposes severe inference bottlenecks, precluding parallelization, and increasing latency at scale, while the rigid causal structure can be suboptimal for tasks requiring bidirectional reasoning and holistic understanding.

Discrete Masked Diffusion Language Models (MDLM) have emerged as a compelling alternative to the prevailing AR paradigm. By reconstructing sequences from random masked inputs, these models inherently support parallel generation and leverage a full bidirectional context, offering a different architectural approach~\citep{diffullama,yu_discrete_2025}. Although these conceptual advantages are clear, the field is still in an early developmental stage. Current research is actively focused on key challenges, including the refinement of specialized training regimes, the design of efficient sampling strategies, the efficient inference of opensource models, and reinforcement learning for MDLM. As a result of this ongoing exploration, most existing diffusion models, including recent advancements like Block Diffusion Language Models (BDLMs)~\citep{Blockdiffusion2025}, operate at a smaller scale (\textit{e.g.}, $\leq$8B parameters). Bridging this scale difference to the hundreds of billions of parameters seen in the leading mainstream AR models is a primary frontier for enabling diffusion models to fully capture complex linguistic patterns for practical deployment.

In this work, we introduce LLaDA2.0 series with 100B/16B total parameters diffusion language models that resolves these fundamental challenges through a novel two-stage continual pre-training (CPT) paradigm. Rather than attempting to train diffusion models from scratch, we leverage existing AR checkpoints as the foundation for a systematic conversion process that preserves linguistic knowledge while introducing diffusion capabilities.

The first stage, CPT aims to transform the foundational AR model into a capable diffusion language model. However, direct conversion is challenging due to the inherent data distribution gap between left-to-right generation and bidirectional denoising. Although the BDLM formulation partially reduces this gap through blockwise masked reconstruction, it suffers from low data utilization, limiting the effective exploitation of large-scale corpora. To this end, we introduce the Warmup–Stable–Decay (WSD) strategy, smoothly bridging the AR-to-dLLM gap while substantially improving CPT efficiency. WSD gradually expands the model’s receptive field to introduce diffusion-style context (\textit{Warmup}), strengthens global denoising under full-sequence training (\textit{Stable}), and then refines the model into an efficient blockwise structure (\textit{Decay}). This progressive adjustment enables a stable and data-efficient transition to diffusion-based learning. Additionally, under full attention in packed training sequences, diffusion models risk forming spurious dependencies across document boundaries, leading to semantic confusion and instability in bidirectional training. To prevent such cross-document interference, we introduce a document-level attention mask that restricts self-attention within individual documents, ensuring coherent context modeling.

The second stage, Post-training for Practical Deployment, transitions the model from a raw predictive engine into a capable and efficient assistant. The random masking nature of the diffusion fine-tuning objective means any single sample provides only a partial learning signal. We address this by employing a complementary masking strategy, which ensures near-100\% data utilization and accelerates convergence by guaranteeing every token contributes to the model's learning. With an efficient foundation for instruction tuning, we then align the model with human preferences by adapting modern techniques like Direct Preference Optimization (DPO)—originally designed for AR models—by reformulating the objective over the model's reconstruction loss. Beyond alignment, practical deployment hinges on inference speed. To realize the full promise of parallel decoding, which is often limited by a model's lack of predictive confidence, we incorporate an auxiliary confidence prediction loss. This trains the model to be ``sharper" and more certain, unlocking aggressive and efficient parallel generation without degrading quality.

We release instruction-tuned variants for practical deployment: \texttt{LLaDA2.0-mini} (16B parameters) for resource-constrained applications and \texttt{LLaDA2.0-flash} (100B parameters) for high-performance scenarios. Both variants retain the parallel decoding advantages of our diffusion training while being optimized for instruction following and safety through comprehensive post-training alignment.

Our contributions provide a practical recipe for the community to leverage AR stability while achieving diffusion parallelism, opening new possibilities for efficient large-scale language modeling.

\section{Related Work}
\subsection{Train dLLMs from scratch}
Auto-regressive language models~\citep{lingv2,moonshot2025kimik2,liu2024deepseekv3,meta2025llama4} are typically trained by maximizing the likelihood of predicting the next token. Under this paradigm, model performance has been shown to scale effectively with increasing model size, dataset volume, and computational resources, following well-established scaling laws. Recently, MDLMs~\citep{song2025seed,Dream7B2025,nie2025llada} have emerged as an alternative generative framework, reformulating text generation as an iterative denoising process. In each forward step, a subset of tokens is randomly masked, and the model is trained to recover the original tokens conditioned on the remaining unmasked context. %

Encouraged by this paradigm shift, several studies have explored training MDLMs from scratch to assess their full potential. For instance, LLaDA~\citep{nie2025llada} demonstrated that a 8B dense MDLM, trained entirely from scratch, achieves performance competitive with similarly sized AR counterparts. Building upon this, LLaDA-MoE~\citep{zhu2025lladamoe} introduced the Mixture-of-Experts (MoE) architecture into the MDLM for the first time, showing that a scratch-trained MoE-based MDLM can surpass dense models in both efficiency and capability, thereby validating the compatibility and scalability of MDLMs with advanced MoE designs. Moreover, due to the fundamentally different training dynamics compared to AR models, established training practices and hyperparameter recipes from the AR domain are often suboptimal for MDLMs. To address this gap, recent efforts such as Quakka~\citep{quakka2025} and OpenMoE2~\citep{openmoe2} have begun investigating the scaling properties and optimal training strategies specifically tailored for MDLMs, laying the groundwork for principled scaling in this emerging paradigm.

However, from-scratch trained MDLMs still lag behind state-of-the-art AR models in overall performance. This gap can be largely attributed to the disparity in training data volume and the maturity of infrastructure support—factors that have been extensively optimized over years of development for AR models. Moreover, due to the high computational cost and long training cycles required for pretraining from scratch, MDLMs mentioned above are typically limited in model scale ($\leq$8B), whereas leading AR models now routinely scale into tens or even hundreds of billions. 

\subsection{Scaling dLLMs with AR initialization}
Given the strong knowledge capacity and performance of AR models, several recent studies have explored initializing dLLMs from pre-trained AR models to reduce training costs and narrow the performance gap between AR models and dLLMs. For instance, DiffusionLLaMA~\citep{diffullama} and Dream-7B~\citep{Dream7B2025} adopt a mask annealing strategy to gradually transition from causal attention to bidirectional attention during training, while employing a CART-based loss reweighting scheme to balance token-level learning dynamics. In contrast, RND1~\citep{RND1} takes a more direct approach by immediately converting the causal attention mechanism of the AR model into a bidirectional one upon initialization. Notably, RND1 observes that when initializing DLM training from an AR model, preserving knowledge-intensive capabilities requires constraining updates to the model’s dense layers to prevent catastrophic forgetting.

Block Diffusion Language Models (BDLMs)~\citep{Blockdiffusion2025} provide a hybrid paradigm that balances efficiency and performance by combining diffusion and AR modeling. Tokens are generated block-wise: within each block, a diffusion process reconstructs masked tokens, while blocks are produced auto-regressively. 
This design enables variable-length generation and supports KV-cache reuse during decoding, enhancing inference efficiency. Consequently, BDLMs can be effectively initialized from AR models, narrowing the performance gap. For example, SDAR~\citep{SDAR2025} leverages the Qwen-3 series~\citep{yang2025qwen3} to train more efficient BDLMs. By exploring various block sizes and optimization strategies, it achieves performance comparable to its AR base model. 

However, one key limitation across all existing methods is their restricted model scale—ranging only from 7B to 30B parameters—leaving the feasibility and scalability of AR-initialized diffusion models largely unexplored at larger scales. 
Besides, the low training efficiency of block diffusion hinders its widely application to large-scale corpus for large-size models. 
Whether such initialization strategies can effectively generalize to models beyond the 30B scale remains an open question.

\subsection{dLLMs post-training}
Beyond pre-training, post-training is crucial for unlocking the full potential of dLLMs by aligning them with specific tasks and human preferences. This process typically involves supervised fine-tuning (SFT) to instill instruction-following capabilities, reinforcement learning (RL) to enhance complex reasoning, and inference optimization to address efficiency bottlenecks.

Recent work has explored SFT to adapt dLLMs for specialized domains. For instance, Dream-Coder~\citep{xie_dream-coder_2025} fine-tunes a 7B dLLM for code generation, demonstrating unique abilities like adaptive "sketch-then-fill" strategies for complex algorithms. Similarly, the general-purpose model Dream-7B~\citep{Dream7B2025} leverages SFT to achieve performance on par with top-tier AR models, while uniquely excelling at tasks requiring complex planning and constraint satisfaction. Other studies have investigated specialized fine-tuning strategies to balance quality and efficiency. Seed-Diffusion~\citep{song2025seed}, for example, employs a two-stage curriculum learning strategy to train a high-speed code generation model, while LiDAR~\citep{liu_tidar_2025} introduces a hybrid "think in diffusion, generate in AR" architecture through fine-tuning, significantly boosting inference throughput while maintaining quality.

To further enhance dLLMs' reasoning abilities, researchers have begun adapting reinforcement learning techniques. However, applying standard policy gradient methods is challenging due to the intractable log-likelihood of dLLMs. To address this, SPG~\citep{wang2025spg} proposes a novel Sandwich Policy Gradient algorithm that obtains a more robust and less biased gradient by maximizing an evidence lower bound for high-reward samples and minimizing an evidence upper bound for low-reward ones. Another line of work, TraceRL~\citep{wang2025revolutionizing}, focuses on aligning the training objective with the model's multi-step generation trajectory. This framework led to the TraDo series of models, which have not only surpassed strong AR models on reasoning benchmarks but also produced the first dLLM capable of long-chain-of-thought reasoning.

A significant challenge for dLLMs is their slow inference speed, stemming from the iterative nature of the denoising process. To mitigate this, several acceleration methods have been proposed. DPad~\citep{chen_dpad_2025} offers a training-free solution by treating future tokens as a dynamic "scratchpad" and using a sliding window and distance-based pruning to reduce redundant computations, achieving a dramatic speedup, especially for long sequence generation. In contrast, D2F~\citep{wang_diffusion_2025} introduces a hybrid autoregressive-diffusion paradigm that enables parallel denoising of future text blocks even before preceding ones are fully generated. This approach allows dLLMs to leverage KV-caching and, for the first time, surpass the inference speed of equivalently sized AR models.

Despite these advances, the field of dLLM post-training is still nascent. Systematic exploration of how these techniques—SFT, RL, and acceleration—interact with one another, and how they scale to models with hundreds of billions of parameters, remains an open and critical area for future research.

\section{LLaDA2.0 Training Paradigm}

\begin{figure}
    \centering
    \includegraphics[width=\linewidth]{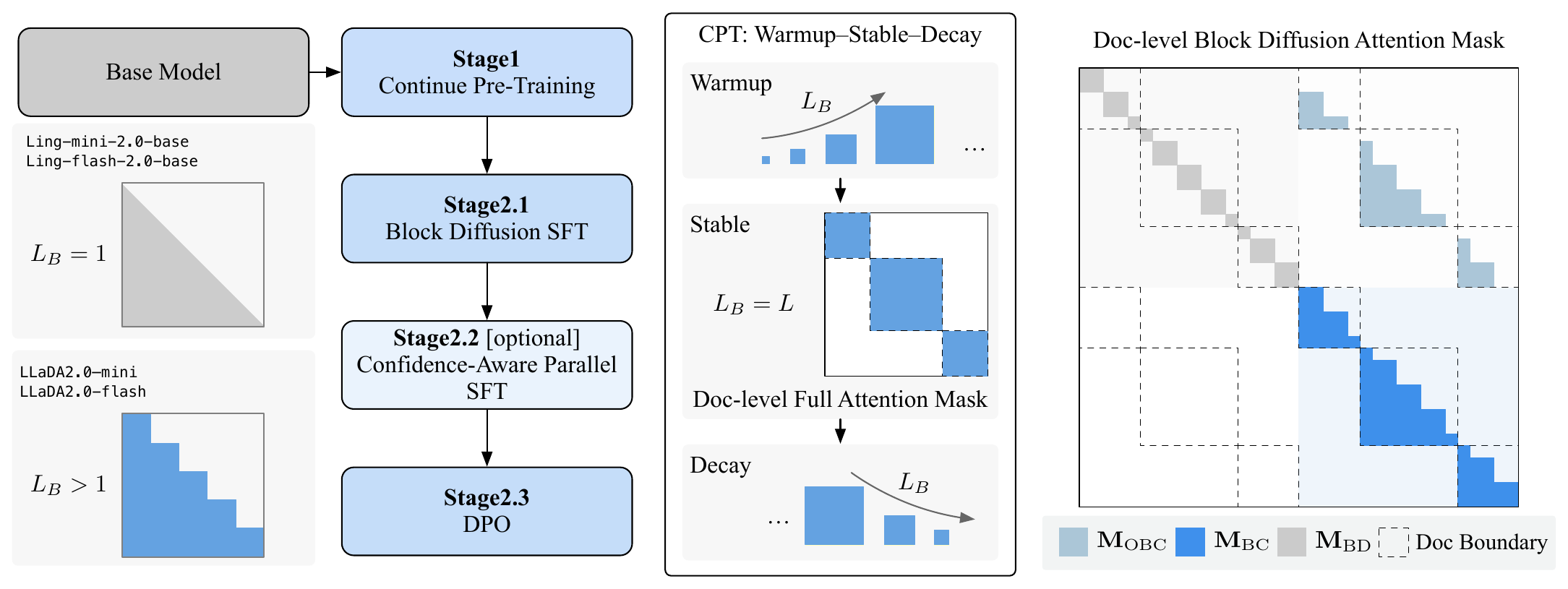}
    \caption{\textbf{A schematic of the progressive training framework for transforming an AR model into a MDLM.} Continual Pre-Training Stage facilitates the \textbf{Warmup-Stable-Decay} strategies by scheduling block size $L_{B}$ enables smooth, stable, and effective attention mask adaptation. Post-training Stage facilitates the same block diffusion configuration conducting the instruction SFT, Confidence-Aware Parallel SFT, and DPO. The right panel illustrates the document-level block diffusion attention mask,which enables an efficient, vectorized forward pass by constructing a single input sequence from multiple noisy and clean examples, such as $[\bm{x}_{\text{noisy1}},\dots,\bm{x}_{\text{clean1}},\dots]$. The forward pass then employs a combination of block-diagonal ($\mathbf{M}_{\text{BD}}$), offset block-causal ($\mathbf{M}_{\text{OBC}}$), and block-causal ($\mathbf{M}_{\text{BC}}$) masks.}
    \label{fig:llada2_paradigm}
\end{figure}

Figure~(\ref{fig:llada2_paradigm}) illustrates the holistic training pipeline of LLaDA2.0, a staged and scalable framework designed to transform AR language models into highly efficient diffusion language models. Our paradigm follows a three-stage progression: (1) \textit{Continual Pre-training} from AR to MDLM, (2) \textit{Block Diffusion Pre-training} to transition from token-level to block-level diffusion modeling, and (3) \textit{Post-training} for alignment and task specialization.

The process begins with a strong AR base model. We first perform continual pre-training to adapt this model into an MDLM, where it learns to reconstruct randomly masked tokens in a bidirectional, denoising fashion. This phase bridges the gap between AR and diffusion-based generation while preserving the representational geometry of the original model.

Building upon the trained MDLM, we then introduce \textit{block diffusion pre-training}, during which the model is further trained to denoise contiguous spans of text—referred to as “blocks”—rather than individual tokens. This shift enables higher computational efficiency and better long-range coherence during generation.

Finally, after mastering non-autoregressive generation at both token and block levels, the model undergoes post-training--including SFT and DPO to align its outputs with human intent, instruction-following capability, and downstream application requirements. This stage ensures that the powerful generative backbone developed during diffusion pre-training translates into practical performance gains across diverse tasks.

Overall, LLaDA2.0’s training paradigm emphasizes \textbf{knowledge inheritance}, \textbf{progressive adaptation}, and \textbf{efficiency-aware design}, enabling seamless evolution from AR models to fluent, flexible, and fast diffusion large language models.

\newtcolorbox{takeaway}[1][]{
  enhanced,
  arc=3mm,
  boxrule=0.5pt,
  colback=myBlueBase!10!white,
  colframe=myBlueBase!80!black,
  sharp corners=south,
  fonttitle=\bfseries,
  coltitle=white, 
before title={\textcolor{yellow!30!white}{\faLightbulb}~~},
  attach boxed title to top left={xshift=2mm, yshift=-2mm},
  boxed title style={
    arc=3mm,
    boxrule=0.3pt, 
    colback=myBlueBase!90!white,
    colframe=myBlueBase!80!black, 
    width=\dimexpr\linewidth-2*0.5pt-2*4mm\relax*0.7, %
  },
  drop shadow={black!20!white},
  #1
}

\section{Continual Pre-training via Warmup-Stable-Decay (WSD)}

\begin{takeaway}[title={Takeaway}]
\begin{itemize}[leftmargin=2em]
    \item[(1)] \textbf{Warmup-Stable-Decay} enables a smooth and data-efficient conversion from AR to dLLMs.
    \item[(2)] The \textbf{document-level attention mask} ensures coherent bidirectional modeling within semantic boundaries.
    \item[(3)] \textbf{Top-k Checkpoint Merge} enhances performance and generalization by averaging the top k model checkpoints.
\end{itemize}
\end{takeaway}

Converting a pre-trained AR language model into a high-performance diffusion language model is fundamentally challenging due to the misalignment in architectural inductive biases and training objectives. While AR models generate tokens sequentially from left to right, diffusion-based models rely on bidirectional context and learn to reconstruct corrupted sequences in arbitrary unmasking orders. A direct objective switch often leads to unstable optimization and severe degradation of pretrained knowledge.

To address this gap, we propose a \textbf{Warmup–Stable–Decay} (WSD) continual pre-training strategy that enables a smooth, stable, and effective transition from AR to dLLM. WSD decomposes the conversion into three coordinated phases: 
\begin{itemize}[leftmargin=2em]
    \item \textbf{Warmup}: Progressively increase the block size in block diffusion language models (BDLM) to gradually transform the AR model into a full-sequence masked diffusion language model (MDLM).
    \item \textbf{Stable}: Stabilize and enrich the model's understanding of diffusion dynamics through large-scale training under the MDLM paradigm.
    \item \textbf{Decay}: Revert back to a compact BDLM with smaller block sizes to achieve better speed-efficiency trade-offs during inference.
\end{itemize}
This progressive schedule preserves the AR model’s priors while steadily adapting it to the structural requirements of diffusion modeling.

Moreover, the \textbf{document-level attention mask} is applied throughout training to all input sequences. This mechanism is crucial for handling packed heterogeneous documents, preventing the model from forming spurious connections across unrelated texts, thereby ensuring semantic coherence and improving learning stability within each document. In addition, we adopt a \textbf{top-k checkpoint merging} strategy~\citep{tian2025wsm}, to enhance generalization by averaging the parameters of the best-performing checkpoints, smoothing the parameter landscape, and yielding a more robust final model with boosted performance.

\subsection{Warmup-Stable-Decay Conversion Strategy}\label{sec:warmup} We begin with the AR base models \texttt{Ling-mini-2.0} and \texttt{Ling-flash-2.0} ~\citep{lingv2}, which can be viewed as a special case of BDLM with block size 1. This perspective allows us to treat AR models as the initial BDLM configuration with minimal granularity.

\paragraph{Phase-1: Progressive Block Size Warmup} The core idea of the warmup phase is to \emph{gradually increase the block size}, thereby expanding the receptive field within which the model performs joint denoising. Starting from block size $L_B = 1$, we incrementally scale it up to 4, 32, then 64, and ultimately reach $L_B = 4096$ — at which point the entire sequence is treated as one single block. To avoid fragmented blocks, we require the sequence length to be divisible by the current block size. At the final enlargement, the BDLM becomes equivalent to a standard MDLM that operates over fully masked sequences with global attention. Crucially, each block-size transition is trained on moderate-scale data to ensure smooth adaptation. This progressive enlargement allows the model to smoothly adapt its internal representations to handle larger contextual spans and more complex masking patterns.

\paragraph{Phase-2: Large Scale Stable Training} Once the block size reaches 4096 and the model transitions to the MDLM pattern, the ``clean'' part of the attention computation (see Figure~\ref{fig:llada2_paradigm}) no longer needs to be maintained. This significantly reduces the computational cost of attention, allowing data to be processed far more efficiently under the MDLM paradigm. With the model now fully adapted to this regime, the stable training phase focuses on deepening its understanding of diffusion dynamics through extensive training on large-scale corpora. At this stage, the block size is fixed at 4096, effectively making every input a single-block sequence, equivalent to the classical MDLM setting.

\paragraph{Phase-3: Block Size Decay} Finally, after large-scale MDLM training, we gradually reduce the block size from 4096 to a small block size (e.g., 32) to convert the model back into an efficient BDLM. This decay process distills the global contextual knowledge learned during MDLM into a compact blockwise structure. By decreasing the block size step-by-step (e.g., starting from 4096 to 2048) rather than abruptly, the model smoothly adapts from global to local conditioning, preserving its semantic understanding while regaining BDLM’s practical benefits such as KV-cache reuse and fast variable-length generation.

\paragraph{Overall Training Objective} The optimization objective of BDLM~\citep{Blockdiffusion2025}  is designed to enable the model to accurately reconstruct the original, uncorrupted tokens within these designated masked blocks using a standard cross-entropy loss. Specifically, we define the training loss during  warmup and decay phases (phase-1\&3) under the BDLM paradigm as:
\begin{equation}\label{eq:bdlm}
    \mathcal{L}_{\text{BDLM}}(\theta) = - \mathbb{E}_{t, \bm{x}_0, \bm{x_t}} \left[ \frac{\alpha_{t}'}{1-\alpha_{t}}\sum_{k=1}^{K} \sum_{i=1}^{L_B} \mathbb{1}[x_{t,k}^i=\text{[MASK]}] \log p_{\theta}(\bm{x}_{0,k}^i | \bm{x}_{0,<k}, \bm{x}_{t,k}) \right],
\end{equation}
where the expectation is over timestep \(t\), the clean sequence \(\bm{x}_0\), and its corrupted version \(\bm{x}_t\) (tokens masked with probability \(1-\alpha_t\)). Indicator $\mathbb{1}[\cdot]$ ensures predictions are made only for masked tokens, and \(-\alpha_t'/(1-\alpha_t)\) is the diffusion‑derived time weight. Here \(K=L_{\text{total}}/L_B\) is the number of blocks, \(L_B\) the block size, \(x_{t,k}^i\) the \(i\)-th token in block \(k\), \(\bm{x}_{0,<k}\) the preceding clean blocks, and \(\bm{x}_{t,k}\) the noisy version of the current block.

During the stable training (phase-2) of MDLM (i.e., K=1), the objective simplifies to:
\begin{equation}\label{eq:mdlm}
    \mathcal{L}_{\text{MDLM}}(\theta) = - \mathbb{E}_{t,\bm{x}_0,\bm{x}_t} [\frac{\alpha'_t}{1-\alpha_t}\sum_{i=1}^{L}\mathbb{1}[x_t^i=\text{[MASK]}]\log p_{\theta}(x^i_0 | \bm{x}_{t})].
\end{equation}

\subsection{Document-level Attention Mask}
Our training sequences are formed by packing heterogeneous documents into fixed-length segments to maximize throughput. However, this introduces artificial long-range dependencies across semantically unrelated texts. Without careful handling, standard attention would incorrectly attend across document boundaries, leading to contextual confusion and significantly hindering the model's ability to perform robust bidirectional modeling crucial for denoising.

To mitigate this fundamental challenge and preserve semantic coherence, we redefine the attention mechanism with a specialized \textbf{block-wise document-level attention mask}. This mask ensures that attention operates strictly within document boundaries, preventing cross-document contamination and allowing the model to fully leverage bidirectional context for accurate reconstruction of corrupted blocks. The native Block Diffusion vectorizes the training process to achieve parallel training of blocks, and this mask is applied accordingly.
Specifically, for a concatenated sequence $x_{full}$ of length $2L$ (comprising $x_t$ followed by $x_0$), and assuming tokens $i$ and $j$ are already confined to the same document segment (as enforced by the initial document-level mask), the attention mask $\bm{M}\in\{0,1\}^{2L\times 2L}$ is constructed by dividing each sequence ($x_t$ and $x_0$) into contiguous blocks. Let $b(k) = \lfloor k / L_B \rfloor$ denote the block index for token $k$ given a block size $L_B$. The mask is defined as:
\begin{equation}
\bm{M}_{ij} =
\begin{cases}
    \mathbb{1}_{b(i) = b(j)} & \text{if } i \in x_t \text{ and } j \in x_t \\
    \mathbb{1}_{b(i) > b(j-L)} & \text{if } i \in x_t \text{ and } j \in x_0 \\
    \mathbb{1}_{b(i-L) \ge b(j-L)} & \text{if } i \in x_0 \text{ and } j \in x_0 \\
    0 & \text{otherwise}
\end{cases}
\end{equation}
Where $i, j \in \{0, 1, \dots, 2L-1\}$ are the indices in the full sequence. The first condition ($\mathbb{1}_{b(i) = b(j)}$) implements block-diagonal attention within the noisy sequence $x_t$. The second ($\mathbb{1}_{b(i) > b(j-L)}$) enables cross-attention from $x_t$ to $x_0$, but only from blocks in $x_t$ to earlier blocks in $x_0$. The third ($\mathbb{1}_{b(i-L) \ge b(j-L)}$) imposes a causal block attention pattern within the clean sequence $x$, allowing a block to attend to itself and all preceding blocks. The "otherwise" condition corresponds to a zero matrix, explicitly preventing attention from queries in $x_0$ to keys in $x_t$. 
This allows each block to leverage context from relevant blocks (according to the mask) for reconstruction, capturing inter-block dependencies while maintaining the causal and block-diagonal principles essential for stable diffusion training. During our exploration, we also experimented with other tricks like random-length~\citep{xie_dream-coder_2025} and CART~\citep{Dream7B2025}. However, the results demonstrate that the document-level attention mask is more fundamental in CPT training compared to these techniques, and it consistently achieves superior performance. As illustrated in Figure~\ref{fig:llada2_paradigm}, this forms a structured attention layout that balances locality and global document coherence.

For MDLM, the document-level attention mask simplifies to $\bm{M}\in\{0,1\}^{L\times L}$, where:
\begin{equation}
\bm{M}_{ij} = 
\begin{cases}
1, & \text{if $i,j$ belong to the same document}, \\
0, & \text{otherwise}.
\end{cases}
\end{equation}

\subsection{Top-k Checkpoint Merge}
To further enhance the generalization and robustness of our Block Diffusion Language Model, we employ a top-k checkpoint merging strategy.
Upon completion of BDLM pre-training, we identify the top $k$ best-performing model checkpoints, typically selected based on validation metrics like perplexity. The parameters (weights and biases) of these $k$ checkpoints are then arithmetically averaged to form a single, unified BDLM. Based on WSM scheduler~\citep{tian2025wsm}, this merge strategy can effectively ensemble diverse ''knowledge" captured by the model at various optimal or near-optimal training states. This smooths the parameter landscape, mitigates overfitting, and yields a more stable and generalizable model. A key advantage of the WSM approach is its optimizer-agnostic nature, allowing seamless integration without altering the underlying training pipeline.
Crucially, this post-training Top-k Merge fundamentally differs from the Exponential Moving Average (EMA). While EMA is an in-training technique that continuously smooths parameters, merging is an offline procedure. It explicitly selects and averages distinct, high-performing model states, consolidating their strengths rather than merely smoothing the final training step.

\section{Post-training}

\begin{takeaway}[title={Takeaway}]
\begin{itemize}[leftmargin=2em]
\item[(1)] Applying \textbf{complementary masking} and a \textbf{mask ratio bandwidth} during SFT improves sample efficiency and stabilizes convergence.
\item[(2)] An auxiliary \textbf{confidence loss} is incorporated to sharpen predictions, which is crucial for efficient parallel decoding.
\item[(3)] \textbf{DPO} is adapted by defining sequence log-probabilities over masked tokens, enabling effective preference alignment for the diffusion model.
\end{itemize}
\end{takeaway}

\subsection{Supervised Fine-Tuning with Block Diffusion}

Following the pre-training phase, the model is aligned to follow user instructions through supervised fine-tuning (SFT). This is achieved by adapting the diffusion training objective to be conditional on an input prompt, $\bm{c}$. The model is thus trained to generate the desired response $\bm{x}_0$ by minimizing the following loss function:
\begin{equation}\label{eq:sft_loss}
    \mathcal{L}_{\text{SFT}}(\theta) = - \mathbb{E}_{t,(\bm{c},\bm{x}_0),\bm{x}_t} \left[\frac{\alpha_{t}'}{1-\alpha_{t}}\sum^{K}_{k=1}\sum_{i=1}^{L_{B}}\mathbb{1}[x_{t,k}^i=\text{[MASK]}]\log p_{\theta}(x^i_{0,k} |\bm{c}, \bm{x}_{0,<k}, \bm{x}_{t,k})\right].
\end{equation}
Here, the model $p_{\theta}$ learns to predict the original tokens $x^i_{0,k}$ of a clean response from a noisy version $\bm{x}_t$. The loss is computed only on masked tokens within the current noisy block $\bm{x}_{t,k}$. To do this, the prediction is conditioned on the prompt $\bm{c}$, auto-regressive context from prior clean blocks $\bm{x}_{0,<k}$, and the current noisy block $\bm{x}_{t,k}$ that it must denoise.

\paragraph{Padding strategies \& Mask ratio bandwidth}
To ensure compatibility with our block-wise attention mask, we quantize each sequence's length. Specifically, the original length is rounded up to the nearest multiple of the block size, $b$. This process defines an "effective length" for each sequence, guaranteeing its boundaries align perfectly with the block boundaries required by the attention mechanism.

To optimize the training dynamics, we further implement a ``mask ratio bandwidth'' strategy. Standard discrete diffusion processes typically sample mask probabilities across the full unit interval, $\alpha_{t} \sim U[0, 1]$. However, as identified by~\cite{Blockdiffusion2025}, extreme masking rates induce high gradient variance while offering minimal learning signal: near-zero masking renders reconstruction trivial, while near-total masking reduces the objective to simply learning data marginals. To mitigate this, we clip the noise schedule, constraining the sampling of mask rates to a bounded interval $[\alpha_{\min}, \alpha_{\max}]$ rather than the full range. This bandwidth restriction focuses the training objective on the noise regimes that provide the most informative gradients, thereby stabilizing convergence and improving the model's generative perplexity.

\paragraph{Complementary Masking}
Complementary Masking~\citep{li_lavida_2025} is a training optimization that enhances the data efficiency of the MDLM objective, $\mathcal{L}_{\text{MDLM}}(\theta)$. The strategy's core principle is to generate two antithetical training instances from a single source sequence $\bm{x}_0$. A primary noised sequence, $\bm{x}_t$, is formed using a random mask, while a complementary sequence, $\bm{x}'_t$, is simultaneously produced using that mask's logical inverse\footnote{We also try complementary masking in CPT and find it only works fine on corpus less than 100B tokens, while it does not show advantages with more training data, so that we only adopt it in post-training.}.

By incorporating both $\bm{x}_t$ and $\bm{x}'_t$ into the same training batch, this method provides a deterministic guarantee: every token position across the sequence length $L$ is presented to the model in its uncorrupted state exactly once within the pair. This not only doubles the effective data utilization from each sample, thereby accelerating convergence, but also entirely eliminates token-level sampling bias. Consequently, the model benefits from a more comprehensive and uniform learning signal at every optimization step, leading to enhanced robustness.

\paragraph{Data Recipe Curation}
A balanced, high-quality SFT dataset underpins the model's capabilities, achieved through a strategic composition of tasks spanning three principal pillars: Reasoning, General, and Industrial. The Reasoning pillar hones analytical and logical faculties through mathematics and code generation. The General pillar cultivates linguistic richness and social intelligence via creative and dialogic tasks. The Industrial pillar embeds domain-specific expertise by simulating end-to-end workflows under real-world constraints. This integrated methodology ensures a holistic skill profile, preventing capability skew and enabling fluid shifts between abstract reasoning and applied problem-solving.

\subsection{Confidence-Aware Parallel Training}

\begin{figure}
    \centering
    \includegraphics[width=0.95\linewidth]{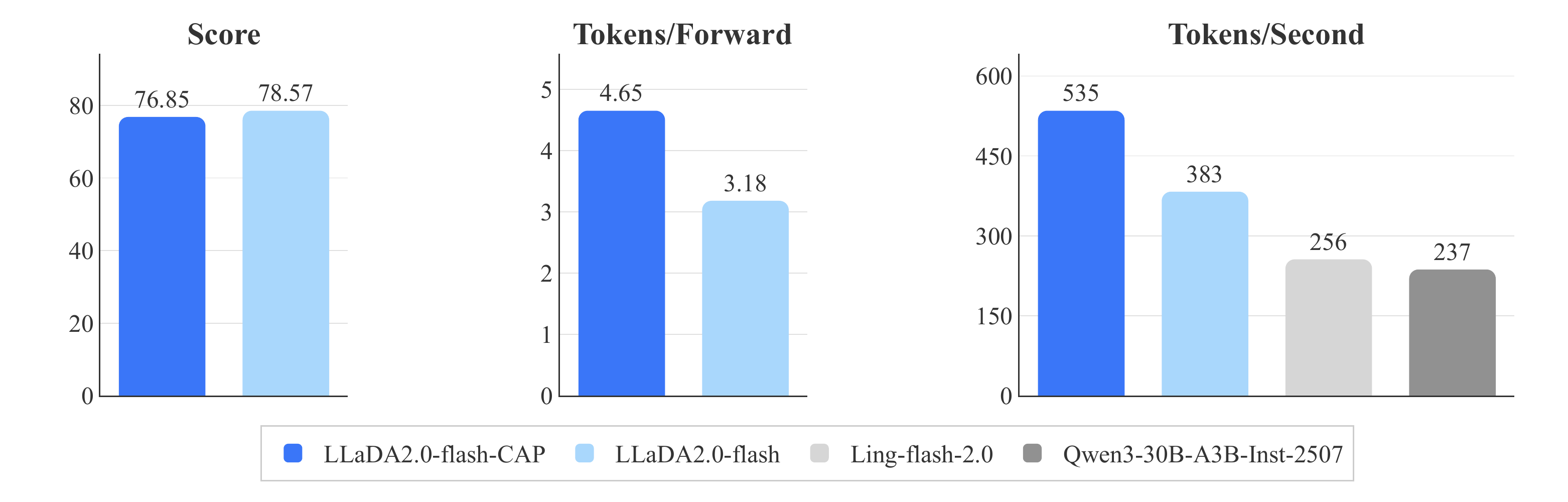}
    \caption{Average score and tokens‑per‑forward (TPF) for LLaDA2.0‑flash with and without CAP across 12 benchmarks. Inference speed (tokens per second) of LLaDA2.0‑flash compared with similarly sized AR models on 4 code and math benchmarks.}
    \label{fig:tps_tparallel}
\end{figure}

To enhance the model's predictive confidence, which is crucial for 
efficient parallel decoding, we propose Confidence-Aware Parallel (CAP) Training. We incorporate an auxiliary confidence 
loss, $\mathcal{L}_{\text{conf}}$, inspired by dParallel~\citep{chen_dparallel_2025}.
The primary objective, $\mathcal{L}_{\text{SFT}}$, ensures correctness but provides diminishing incentive to sharpen the predictive distribution for tokens that are already correctly predicted. The confidence loss addresses this by selectively minimizing the entropy of the model's output distribution, $p_{\theta}(\bm{x}_0 | \bm{x}_t, \bm{c})$, but only for the subset of tokens that are correctly predicted in a given step. This compels the model to increase its certainty on its correct predictions.
The final training objective is a weighted combination of the two losses:
\begin{equation}\label{eq:dparallel}
\mathcal{L}(\theta) = \mathcal{L}_{\text{SFT}}(\theta) + \lambda \mathcal{L}_{\text{conf}}(\theta),
\end{equation}
where $\lambda$ is a hyperparameter that balances the two objectives. As illustrated in Figure~\ref{fig:tps_tparallel}, CAP training effectively improves the decoding efficiency of LLaDA2.0-flash while maintaining competitive compression performance, demonstrating a favorable trade-off between generation quality and inference speed.

\subsection{DPO}

Building upon the SFT stage, we further align the policy model $\pi_{\theta}$ with human intent using Direct Preference Optimization. To support this, we constructed a comprehensive dataset comprising 1.5 million preference pairs across diverse domains, including general knowledge, mathematics, and instruction following. To ensure a stable transition in optimization, the learning rate for the DPO stage is initialized consistently with the final learning rate of the preceding SFT phase.

Since the policy model $\pi_{\theta}$ is trained to reconstruct clean tokens $\bm{x}_0$ from noisy blocks $\bm{x}_t$ conditioned on context $\bm{c}$, the standard DPO formulation—which requires exact log-likelihoods—is intractable. Following established practices for diffusion models, we substitute the conditional log-likelihoods with their ELBO. 
We first define the conditional Block Diffusion ELBO, $B_{\text{BDLM}}(\theta, \boldsymbol{x} | \boldsymbol{c})$, for a response $\boldsymbol{x}$. This term mirrors the inner objective of our SFT loss (\eqref{eq:sft_loss}) and is estimated via a single Monte Carlo sample over timesteps and noise:
\begin{equation}\label{eq:bdlm_elbo}
    B_{\text{BDLM}}(\theta, \boldsymbol{x} | \boldsymbol{c}) = \mathbb{E}_{t, \bm{x}_t} \left[ \frac{\alpha_{t}'}{1-\alpha_{t}}\sum_{k=1}^{K} \sum_{i=1}^{L_B} \mathbb{1}[x_{t,k}^i=\text{[MASK]}] \log p_{\theta}(x^i_{k} | \boldsymbol{c}, \boldsymbol{x}_{<k}, \boldsymbol{x}_{t,k}) \right].
\end{equation}
Given a preference pair $(\boldsymbol{x}_w, \boldsymbol{x}_l)$, where $\boldsymbol{x}_w$ is the preferred response and $\boldsymbol{x}_l$ is the dispreferred response, the DPO objective maximizes the margin between the ELBO estimates of the policy $\pi_{\theta}$ and the frozen reference model $\pi_{\theta_{\text{ref}}}$ (initialized from the post-SFT model). The final loss function is defined as:
\begin{equation}\label{eq:dpo_loss}
\mathcal{L}_{\text{DPO}}(\theta) = - \mathbb{E}_{(\boldsymbol{c}, \boldsymbol{x}_w, \boldsymbol{x}_l) \sim \mathcal{D}} \left[ \log \sigma \left( \beta \left[ \Delta B(\boldsymbol{x}_w | \boldsymbol{c}) - \Delta B(\boldsymbol{x}_l | \boldsymbol{c}) \right] \right) \right],
\end{equation}
where $\Delta B(\boldsymbol{x} | \boldsymbol{c}) = B_{\text{BDLM}}(\theta, \boldsymbol{x} | \boldsymbol{c}) - B_{\text{BDLM}}(\theta_{\text{ref}}, \boldsymbol{x} | \boldsymbol{c})$ represents the ELBO advantage of the policy over the reference model, and $\beta$ is a hyperparameter (set to 0.1) that controls the deviation from the reference policy.

\subsection{Inference}
We sample one block at a diffusion step, conditioned on previously sampled blocks $p_{\theta}(\bm{x}^{b}_{s}|\bm{c},\bm{x}^{<b}_{t})$. The generation of each block is itself a multi-step iterative refinement process. At each step, candidate tokens are sampled for all remaining unfilled positions within the block. A hybrid acceptance strategy is then employed: we first accept all tokens whose sampling probability exceeds a predefined confidence `threshold`. If an insufficient number of tokens meet this criterion, a low-confidence fallback is triggered, where we instead accept a fixed number of the most probable tokens regardless of their absolute confidence. This dual mechanism ensures steady generation progress.

\renewcommand{\thesubfigure}{\alph{subfigure}}
\section{Evaluation}
\subsection{Setup}
To comprehensively evaluate the quality of instruction-tuned models, we employ a diverse suite of benchmarks categorized into five dimensions:

\begin{itemize}
    \item \textbf{Knowledge}: MMLU \citep{hendrycks2020mmlu}, MMLU-Pro \citep{wang2024mmlupro}, GPQA-Diamond \citep{rein2024gpqa}, ARC \citep{clark2018think}, CMMLU \citep{li2023cmmlu} C-Eval \citep{huang2023ceval}, GAOKAO-Bench \citep{zhang2023evaluating}, SciBench \citep{wang2023scibench}, PHYBench \citep{qiu2025phybench}, TriviaQA \citep{joshi2017triviaqa}
    \item \textbf{Reasoning}: SQuAD 2.0 \citep{rajpurkar2018know}, DROP \citep{dua2019drop}, KOR-Bench \citep{ma2024kor}, HellaSwag \citep{zellers2019hellaswag}, BIG-Bench Hard \citep{suzgun2023challenging}, BIG-Bench Extra Hard \citep{kazemi2025big}, MuSR \citep{sprague2023musr}, ZebraLogic \citep{lin2025zebralogic}, PrOntoQA \citep{saparov2022language}, PIQA \citep{bisk2020piqa}, OCNLI \citep{hu2020ocnli}, BIG-Bench Hard-CN \citep{opencompass}
    \item \textbf{Coding}: CRUXEval \citep{gu2024cruxeval}, MBPP \citep{austin2021program}, MultiPL-E \citep{cassano2023multiple}, HumanEval \citep{chen2021evaluating}, BigCodeBench \citep{zhuo2024bigcodebench}, LiveCodeBench \citep{jain2024livecodebench}, Spider \citep{yu2018spider}, BIRD \citep{li2023can}, HumanEval+ \citep{liu2023your}, MBPP+ \citep{liu2023your}, HumanEvalFix \citep{muennighoff2023octopack}, Aider \citep{aider}, HumanEval-CN \citep{opencompass}
    \item \textbf{Math}: GSM8K \citep{cobbe2021gsm8k}, MATH \citep{hendrycks2021math}, OlympiadBench \citep{he2024olympiadbench}, AIME 2025 \citep{aime2025aime}, Omni-MATH \citep{gao2024omni}, HARDMath2 \citep{roggeveen2025hardmath2}, GSM-Plus \citep{li2024gsm}, CMATH \citep{wei2023cmath}
    \item \textbf{Agent \& Alignment}: BFCL \citep{patil2025bfcl}, IFEval \citep{zhou2023ifeval}, CodeIF-Bench \citep{wang2025codeif}, Nexus Function Calling Benchmark \citep{nexusraven}
\end{itemize}

This extensive evaluation suite, comprising a total of 47 benchmarks, provides a holistic foundation for assessing model capabilities. In our experiments, we compare the LLaDA2.0 series against strong open-source auto-regressive (AR) models.

For all LLaDA2.0 models, we utilize a temperature of 0.0, a block size of 32, and a decoding threshold of 0.95.

\subsection{Results}
The overall results, presented in the following tables, indicate that the LLaDA2.0 architecture is not only highly competitive, but also shows a promising trend of closing the performance gap with, and even surpassing, AR models in specific key areas. Our models consistently demonstrate strong, and often superior, performance in complex, structured tasks. For instance, LLaDA2.0-mini already outperforms a comparable AR model (Qwen3-8B) in the domains of Reasoning, Coding, and Math. This signal is amplified in our larger model, as LLaDA2.0-flash achieves parity with the powerful Qwen3-30B-A3B-Instruct-2507 and establishes a lead in the critical Coding and Agent domains. This suggests that as diffusion models scale, their inherent strengths in structured generation and tool use become increasingly apparent.

\begin{table}[htbp]
\centering
\caption{Benchmark Performance of LLaDA2.0-mini}
\label{tab:mini_benchmark}
\resizebox{\textwidth}{!}{%
\begin{tabular}{lcccc}
\toprule
\textbf{Benchmark} & \textbf{Qwen3-8B (no\_think)} & \textbf{Ling-mini-2.0} & \textbf{LLaDA2.0-mini-preview} & \textbf{LLaDA2.0-mini} \\
\midrule
\textbf{Average} & 63.42 & 65.77 & 54.67 & 64.34 \\
\midrule
\multicolumn{5}{c}{\textbf{Knowledge}} \\
\midrule
MMLU & 80.94 & 82.15 & 72.49 & 80.53 \\
MMLU-Pro & 65.48 & 63.72 & 49.22 & 63.22 \\
CMMLU & 79.17 & 80.84 & 67.53 & 79.50 \\
C-EVAL & 81.36 & 82.10 & 66.54 & 81.38 \\
GAOKAO-Bench & 84.94 & 87.23 & 74.46 & 84.30 \\
ARC-c & 93.35 & 93.09 & 89.15 & 93.56 \\
GPQA & 46.59 & 56.80 & 23.74 & 47.98 \\
SciBench & 2.85 & 5.28 & 4.10 & 3.53 \\
PHYBench & 9.76 & 14.59 & 5.08 & 11.70 \\
TriviaQA & 52.51 & 55.63 & 50.49 & 51.33 \\
\midrule
\multicolumn{5}{c}{\textbf{Reasoning}} \\
\midrule
BIG-Bench Hard & 79.48 & 83.70 & 70.64 & 78.21 \\
BIG-Bench Extra Hard & 18.27 & 14.81 & 12.36 & 16.47 \\
bbh-zh & 80.09 & 66.11 & 66.62 & 75.75 \\
MuSR & 70.02 & 71.36 & 56.77 & 71.48 \\
ZebraLogic & 37.48 & 79.85 & 14.80 & 64.20 \\
PrOntoQA & 93.12 & 96.06 & 70.00 & 86.00 \\
PIQA & 88.30 & 87.54 & 84.33 & 86.51 \\
OCNLI & 61.49 & 60.17 & 58.68 & 64.51 \\
HellaSwag & 79.56 & 69.02 & 74.01 & 79.01 \\
KOR-Bench & 54.48 & 62.72 & 37.26 & 50.40 \\
DROP & 84.56 & 78.80 & 79.49 & 81.91 \\
SQuAD 2.0 & 85.21 & 75.56 & 85.61 & 86.50 \\
\midrule
\multicolumn{5}{c}{\textbf{Coding}} \\
\midrule
CRUXEval-O & 74.06 & 76.12 & 61.88 & 71.62 \\
MBPP & 78.92 & 84.07 & 77.75 & 81.50 \\
MBPP+ & 71.96 & 76.46 & 66.67 & 74.07 \\
MultiPL-E & 61.70 & 67.09 & 62.43 & 67.46 \\
HumanEval & 84.76 & 85.98 & 80.49 & 86.59 \\
HumanEval+ & 78.66 & 81.71 & 71.95 & 79.88 \\
HumanEvalFix & 76.02 & 82.83 & 60.16 & 74.90 \\
HumanEval-cn & 74.39 & 71.34 & 73.17 & 78.66 \\
BigCodeBench-Full & 36.05 & 35.00 & 30.44 & 32.89 \\
LiveCodeBench & 26.38 & 34.97 & 19.82 & 31.50 \\
Aider & 55.64 & 49.62 & 28.57 & 39.85 \\
BIRD-SQL & 36.11 & 39.67 & 27.71 & 39.34 \\
Spider & 72.80 & 76.43 & 75.64 & 76.76 \\
\midrule
\multicolumn{5}{c}{\textbf{Math}} \\
\midrule
GSM8K & 93.63 & 94.62 & 89.01 & 94.24 \\
MATH & 86.28 & 94.66 & 73.50 & 93.22 \\
OlympiadBench & 55.33 & 72.30 & 36.30 & 67.70 \\
AIME 2025 & 22.08 & 47.66 & 10.00 & 36.67 \\
HARDMath2 & 7.58 & 9.95 & 0.95 & 0.47 \\
Omni-MATH & 33.20 & 48.80 & 19.20 & 41.70 \\
GSM-Plus & 86.09 & 87.82 & 81.44 & 86.24 \\
CMATH & 95.42 & 96.40 & 90.53 & 95.72 \\
\midrule
\multicolumn{5}{c}{\textbf{Agent \& Alignment}} \\
\midrule
IFEval-strict-prompt & 86.90 & 76.16 & 62.50 & 80.78 \\
BFCL v3 & 70.08 & 53.98 & 74.11 & 70.90 \\
CodeIF-Bench & 50.00 & 46.00 & 48.00 & 48.00 \\
Nexus FC & 37.71 & 34.38 & 33.68 & 35.18 \\
\bottomrule
\end{tabular}%
} 
\end{table}
\begin{table}[htbp]
\centering
\caption{Benchmark Performance of LLaDA2.0-flash}
\label{tab:flash_benchmark}
\resizebox{\textwidth}{!}{%
\begin{tabular}{lcccc}
\toprule
\textbf{Benchmark} & \textbf{Qwen3-30B-A3B-Instruct-2507} & \textbf{Ling-flash-2.0} & \textbf{LLaDA2.0-flash-preview} & \textbf{LLaDA2.0-flash} \\
\midrule
\textbf{Average} & 73.60 & 72.15 & 65.97 & 73.18 \\
\midrule
\multicolumn{5}{c}{\textbf{Knowledge}} \\
\midrule
MMLU & 87.13 & 87.98 & 83.15 & 87.69 \\
MMLU-Pro & 74.23 & 76.84 & 66.16 & 73.36 \\
CMMLU & 86.36 & 86.59 & 79.64 & 85.13 \\
C-EVAL & 88.17 & 88.03 & 79.28 & 86.75 \\
GAOKAO-Bench & 94.53 & 93.24 & 86.12 & 93.90 \\
ARC-c & 95.81 & 95.08 & 93.90 & 95.93 \\
GPQA & 57.34 & 67.12 & 41.92 & 61.98 \\
SciBench & 4.54 & 4.14 & 5.13 & 4.13 \\
PHYBench & 29.84 & 27.67 & 7.58 & 30.06 \\
TriviaQA & 65.61 & 69.76 & 69.25 & 66.88 \\
\midrule
\multicolumn{5}{c}{\textbf{Reasoning}} \\
\midrule
BIG-Bench Hard & 85.54 & 89.36 & 82.85 & 86.75 \\
BIG-Bench Extra Hard & 37.80 & 23.24 & 16.70 & 27.86 \\
BIG-Bench Hard - CN & 86.18 & 75.09 & 83.38 & 87.52 \\
MuSR & 79.15 & 82.72 & 78.75 & 80.48 \\
ZebraLogic & 90.97 & 87.60 & 39.90 & 82.30 \\
PrOntoQA & 97.12 & 97.88 & 93.50 & 96.50 \\
PIQA & 91.57 & 91.95 & 91.84 & 92.76 \\
OCNLI & 71.59 & 65.36 & 69.39 & 71.63 \\
HellaSwag & 86.31 & 81.59 & 86.00 & 84.97 \\
KOR-Bench & 68.00 & 68.96 & 53.28 & 64.24 \\
DROP & 87.57 & 88.32 & 88.17 & 87.90 \\
SQuAD 2.0 & 89.51 & 81.32 & 90.61 & 90.00 \\
\midrule
\multicolumn{5}{c}{\textbf{Coding}} \\
\midrule
CRUXEval-O & 86.75 & 82.75 & 74.50 & 85.12 \\
MBPP & 86.65 & 85.01 & 86.65 & 88.29 \\
MBPP+ & 78.04 & 76.19 & 75.93 & 79.63 \\
MultiPL-E & 70.67 & 65.76 & 72.38 & 74.87 \\
HumanEval & 93.29 & 85.98 & 88.41 & 94.51 \\
HumanEval+ & 88.41 & 85.98 & 82.32 & 87.80 \\
HumanEvalFix & 91.16 & 92.68 & 83.33 & 90.24 \\
HumanEval-CN & 87.20 & 74.39 & 84.76 & 89.02 \\
Bigcodebench-Full & 41.49 & 40.70 & 40.44 & 41.58 \\
LiveCodeBench & 41.63 & 44.11 & 29.07 & 42.29 \\
Aider & 71.43 & 71.43 & 51.13 & 66.92 \\
Spider & 81.79 & 80.58 & 81.37 & 82.49 \\
BIRD-SQL & 47.75 & 47.49 & 45.34 & 45.76 \\
\midrule
\multicolumn{5}{c}{\textbf{Math}} \\
\midrule
GSM8K & 96.36 & 95.45 & 95.75 & 96.06 \\
MATH & 96.70 & 96.10 & 83.52 & 95.44 \\
OlympiadBench & 77.59 & 76.19 & 49.33 & 74.07 \\
AIME 2025 & 61.88 & 55.89 & 23.33 & 60.00 \\
HARDMath2 & 4.27 & 23.70 & 3.79 & 4.27 \\
Omni-MATH & 54.00 & 53.00 & 24.60 & 50.30 \\
GSM-Plus & 89.45 & 89.83 & 88.25 & 89.64 \\
CMATH & 96.58 & 96.52 & 95.26 & 96.90 \\
\midrule
\multicolumn{5}{c}{\textbf{Agent \& Alignment}} \\
\midrule
IFEval-strict -prompt & 84.29 & 81.52 & 75.60 & 81.70 \\
BFCL v3 & 73.19 & 67.57 & 74.86 & 75.43 \\
CodeIF-Bench & 54.00 & 56.00 & 56.00 & 58.00 \\
Nexus FC & 49.93 & 36.25 & 47.98 & 50.45 \\
\bottomrule
\end{tabular}%
} 
\end{table}

As shown in \Cref{tab:mini_benchmark}, \textbf{LLaDA2.0-mini} achieves a competitive average score of 64.34, closely approaching its AR peer, Ling-mini-2.0 (65.77). This demonstrates the fundamental viability of the diffusion approach. More importantly, it shows promising signals in complex tasks, outperforming its direct competitor on reasoning benchmarks like SQuAD 2.0 (86.50) and demonstrating more robust instruction following on IFEval (80.78). Its strong performance in coding tasks such as HumanEval (86.59) further suggests an early aptitude for structured generation.

This potential becomes even more evident with our larger model, \textbf{LLaDA2.0-flash}. As shown in \Cref{tab:flash_benchmark}, with an average score of 73.18, it stands firmly on par with strong AR models such as Qwen3-30B-A3B-Instruct-2507 (73.60). Crucially, LLaDA2.0-flash begins to exhibit clear advantages in complex generative tasks, a sign that the diffusion architecture may hold inherent strengths. In the critical domain of coding, it consistently outperforms its AR peers, scoring higher on HumanEval (94.51), MBPP (88.29) and MultiPL-E (74.87). This trend of surpassing AR models also extends to agent capabilities (BFCL v3: 75.43) and advanced mathematics (AIME 2025: 60.00).

In conclusion, the LLaDA2.0 series successfully demonstrates that diffusion-based language models are a powerful and scalable alternative to the dominant auto-regressive paradigm. While rapidly narrowing the gap on general benchmarks, they are already showcasing the potential to surpass traditional architectures in complex, structured domains like code generation and tool use. This positions diffusion models as a highly promising direction for the future of language generation.

\subsection{Analysis}
\paragraph{Analysis of Inference Hyper-parameters}

In addition to our main evaluation, we conducted a brief analysis to tune key inference hyperparameters. To ensure efficiency, this analysis was performed on our LLaDA2.0-mini model, using a representative subset of our benchmarks to understand the trade-off between generation quality (score) and inference speed (measured as TPF - Tokens Per Forward; higher is faster).

\begin{figure}[t]
    \centering
    \small
    \begin{minipage}{0.47\textwidth}
        \centering
        \includegraphics[width=\linewidth]{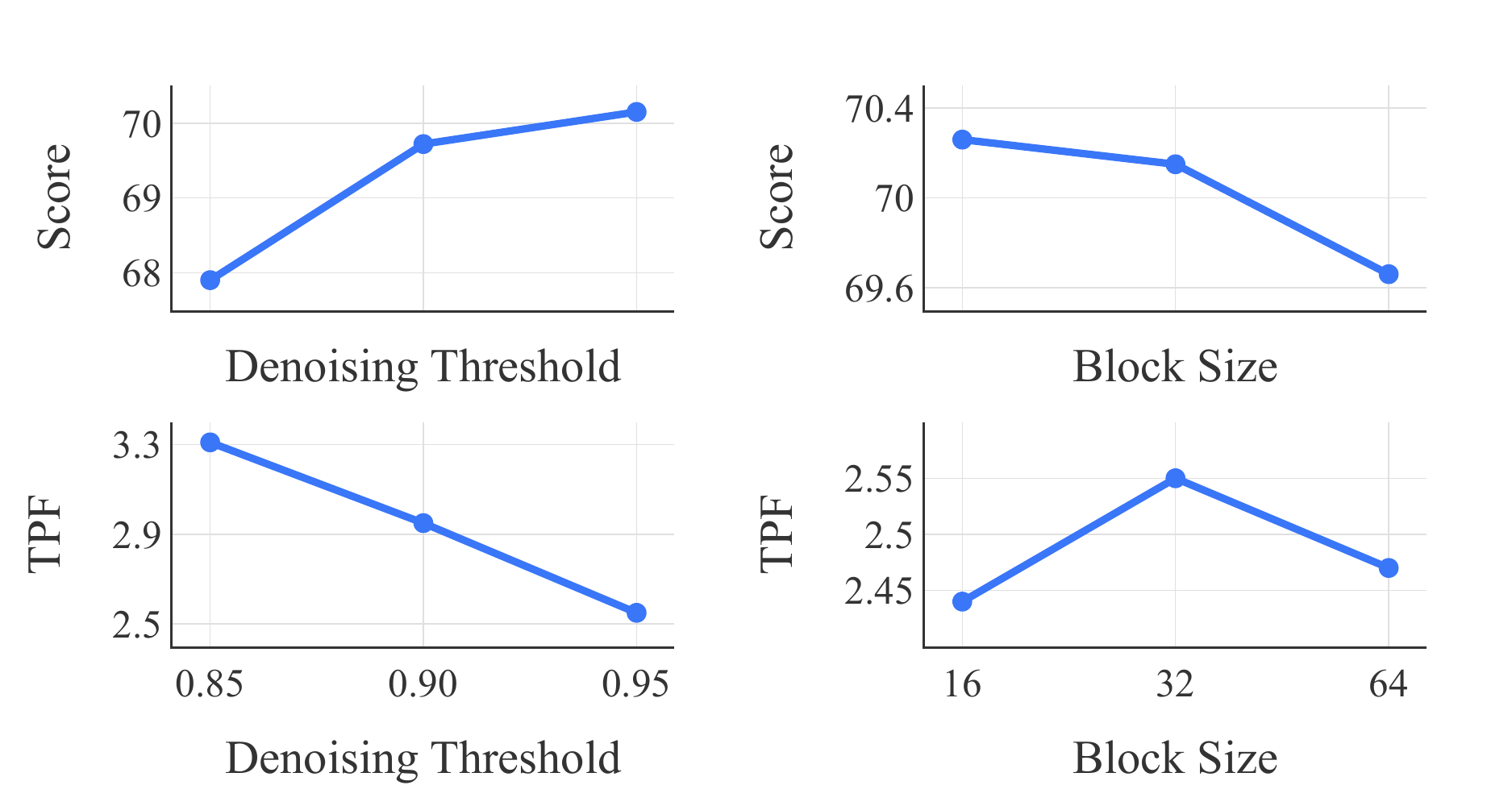}
        \caption{Score/TPF vs threshold/block size}
        \label{fig:params}
    \end{minipage}%
    \hfill
    \begin{minipage}{0.5\textwidth}
        \centering
        \includegraphics[width=\linewidth]{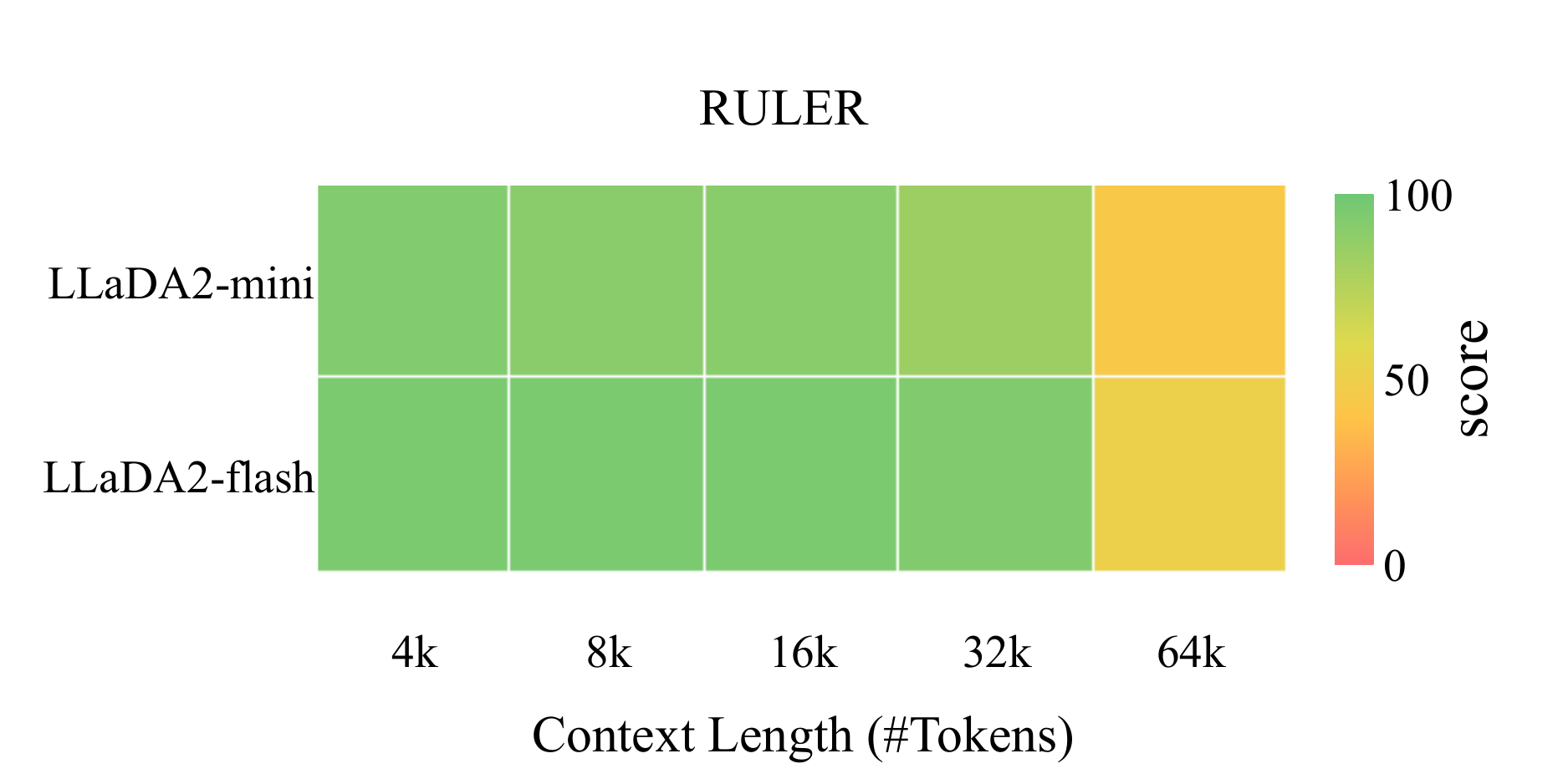}
        \caption{Performance on the RULER benchmark.}
        \label{fig:ctxlen}
    \end{minipage}
    \vspace{-2ex}
\end{figure}

\textbf{Denoising Threshold}. We first investigate the impact of the Denoising Threshold. While keeping the Block Size fixed at 32, we varied the threshold and observed its effect on quality and speed. As shown in \Cref{fig:params}, the results reveal a clear trade-off. A threshold of 0.95 achieved the highest quality score (70.15) at the cost of the lowest inference speed (2.55 TPF). Lowering the threshold to 0.85 boosted the speed to its peak (3.31 TPF), but led to an unacceptable degradation in quality, with the score dropping to 67.90.

\textbf{Block Size}. Subsequently, we analyze the effect of Block Size. We set the Denoising Threshold to 0.95, the optimal value identified in the prior experiment. The results in \Cref{fig:params} demonstrate a similar trade-off. A block size of 16 yielded the highest score (70.26) but with the slowest inference (2.44 TPF). In contrast, increasing the block size to 32 substantially improved the speed to 2.55 TPF with only a marginal quality drop to 70.15. Further increasing the block size to 64 proved suboptimal, as it degraded both score and speed relative to the size-32 setting. Therefore, a block size of 32 emerges as the most compelling choice, offering a significant speed-up for a negligible performance cost.

\textit{In summary}, based on this analysis, the configuration for our main evaluation is well-supported. The Denoising Threshold of 0.95 is the clear choice for maximizing quality. For block-size, the setting of 32 represents an optimal balance, providing the highest throughput with virtually no sacrifice in performance compared to the slightly higher-scoring but slower setting of 16.

\paragraph{Analysis of Context Length}

To rigorously validate our model's performance across various context lengths, we conducted a series of evaluations using the RULER benchmark.

As shown in \Cref{fig:ctxlen}, both models demonstrate strong performance and stability within context length of 32k. The LLaDA2.0-flash model is particularly robust, maintaining a score above 93 across all lengths from 4k to 32k. The LLaDA2.0-mini model also achieves high scores, starting at 93.29 for 4k but showing a degradation to 83.94 at 32k.

To test the models' extrapolation capabilities, we extended the context length to 64k. This was achieved by employing dynamic RoPE scaling during inference, specifically using the YaRN method with a scaling factor of 2.0. However, this extension resulted in a performance degradation for both models, demonstrating a clear trade-off between context length extension and task accuracy.

\textit{In summary}, this evaluation highlights two key findings: (1) The LLaDA2.0 models are exceptionally robust for long-context tasks within their native 32k window. (2) They can be successfully extended to handle 64k sequences via YaRN scaling, providing flexibility for extreme-length applications, albeit with a predictable performance cost.
\section{Training \& Inference Infrastructure}

\subsection{Pretraining}

\begin{wrapfigure}{r}{0.5\textwidth}
    \centering
    \vspace{-1pt}  %
    \includegraphics[width=0.49\textwidth]{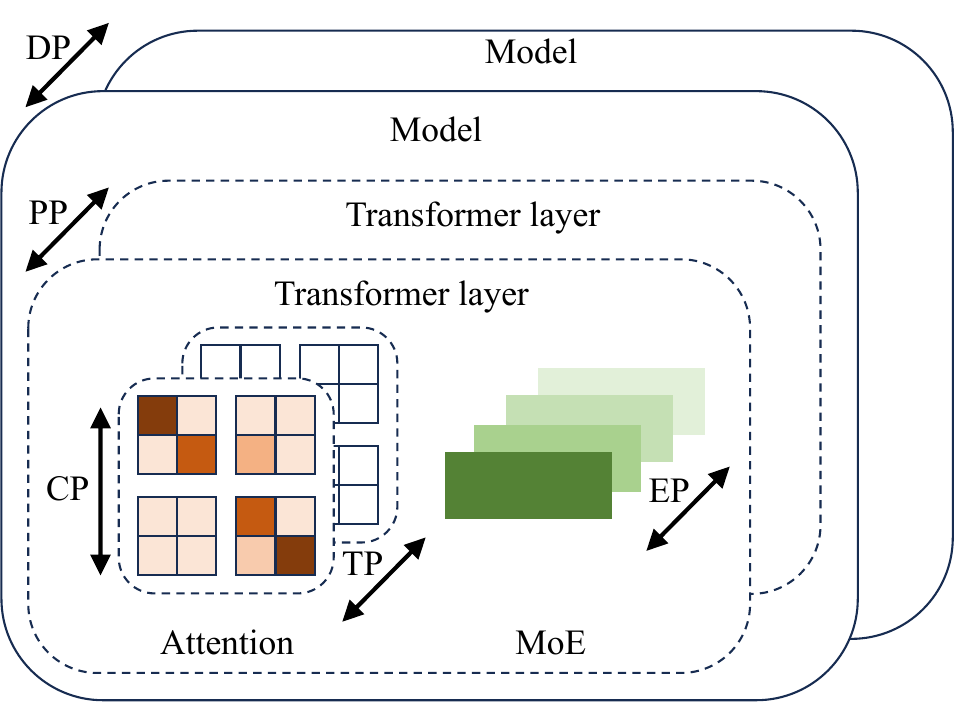}
    \caption{Parallelism overview.}
    \label{fig:megatron_parallelism}
    \vspace{-2ex}  %
\end{wrapfigure}

We adopt Megatron-LM~\citep{megatron-lm} as the pretraining backend to enable efficient training of a 100B-parameter model with long sequences, leveraging data parallelism (DP), pipeline parallelism (PP), tensor parallelism (TP), context parallelism (CP), and expert parallelism (EP), as Figure \ref{fig:megatron_parallelism} shows. To ensure consistency of masked tokens, we generate masked tokens on a single model-parallel (MP, that is, TP and PP) rank and then broadcast to all other ranks within the MP ranks.

\paragraph{Efficient Block Diffusion Training} For flexible support of arbitrary block diffusion attention mask, we utilize cuDNN as the backend for the attention mechanism. This approach achieves more than 1.3x end-to-end speedup and over 90\% memory savings in the attention layer compared to the unfused attention implementation in TransformerEngine when training LLaDA2.0-mini. We further apply a zig-zag partitioning strategy to the block diffusion attention mask to achieve effective load balancing across the CP group.

\paragraph{Numerical Stability} During the transition from AR to diffusion models, training can suffer from gradient explosion, especially at high mask ratios within a document. This issue stems from the fact that masked token embeddings are set to zero during AR training, as these tokens are never observed, leading their corresponding weights to gradually decay to zero. A straightforward fix, randomly reinitializing the masked token embeddings upon loading the AR model, may disrupt other well-trained parameters, potentially causing catastrophic forgetting. To mitigate this while preserving pre-trained knowledge, we instead add independent Gaussian noise to the output of the embedding layer for each masked token during the initial iterations of training. This ensures that the L2 norm of the masked token’s embedding remains significant to avoid gradient explosion, thereby stabilizing the training process.

\subsection{Post-Training}
For the post-training phase, we leverage dFactory\footnote{https://github.com/inclusionAI/dFactory}~\citep{dfactory}, a repository providing efficient training recipes for dLLMs. Built upon the VeOmni~\citep{ma2025veomni} distributed training framework, dFactory allows us to effectively implement complex parallelization schemes. Specifically, our setup for fine-tuning LLaDA2.0 combines Data Parallelism (DP) and Expert Parallelism (EP) to ensure scalable and stable training. To further enhance data throughput and hardware utilization, we adopt a data packing strategy analogous to those used in continued pre-training, which concatenates multiple short sequences into a single longer sequence. This integrated approach provides a robust and high-performance infrastructure for the post-training of our model.

\subsection{Inference Engine}
We adapt dInfer\footnote{https://github.com/inclusionAI/dInfer}~\citep{dinfer}--originally built for high-performance diffusion LLM inference--to efficiently support block diffusion inference. This requires the inference engine to leverage optimization techniques traditionally designed for AR models. For instance, the framework can now effectively exploit KV-cache reuse to substantially reduce prefill computation.
As block diffusion inference closely resembles auto-regressive generation in the execution pattern, we also incorporated block diffusion inference support into SGLang\footnote{https://github.com/sgl-project/sglang/issues/12766}~\citep{sglang}, allowing it to benefit from the same class of system-level optimizations designed for AR models. More mature features in dInfer are undergoing to transport to SGLang.

\paragraph{Inference speed}

\Cref{fig:tps_tparallel} compares the average inference throughput (Tokens Per Second, TPS = \#decoding tokens/\#total-time) of our optimized LLaDA2.0-flash models against state-of-the-art AR models of similar scale on four reasoning and code-generation benchmarks (HumanEval, MBPP, GSM8K, and CRUXEval). All models are evaluated under a consistent generation setup. For diffusion-based models (LLaDA2.0-flash and LLaDA2.0-flash-CAP), we adopt a threshold decoder with a threshold of 0.95. The AR baselines (Ling-flash-2.0 and Qwen3-30B-A3B-Instruct-2507) are deployed using SGLang, while the diffusion models are served with dInfer, ensuring fair performance comparison in real inference environments.
As shown, LLaDA2.0-flash-CAP reaches 535 TPS, outperforming the standard LLaDA2.0-flash (383 TPS) and providing up to 2.1× speed-up over the AR baselines (256 TPS and 237 TPS).

\section{Conclusion}
In this work, we introduced LLaDA2.0, discrete diffusion language models scaling up to 100B total parameters through systematic conversion from auto-regressive models, as well as a set of novel and comprehensive recipes designed to smooth and effectively transform traditional AR language models into highly efficient and performant Masked Diffusion Language Models. 

Through extensive evaluations, it validates the feasibility of the training paradigm. The LLaDA2.0-mini and LLaDA2.0-flash models achieve performances that are competitive with their AR counterparts.
Slightly surprisingly, LLaDA2.0-flash seems to have demonstrated advantages in complex, structured domains such as code generation, mathematical reasoning, and agentic tool use. 
These may have opened a new door to future work in the agentic LLM era while solidifying a gaugeable potential of dLLM for test-time scaling.

Future work may point to further scaling of the parameter volume, RL/thinking paradigm and extending the decoding speed to its extreme.

\bibliographystyle{antgroup}

\bibliography{ref/reference}

\clearpage

\end{document}